\documentclass[letterpaper]{article} 
\usepackage{aaai2027}  
\usepackage[hyphens]{url}  
\usepackage{graphicx} 
\usepackage{natbib}  
\usepackage{caption} 
\nocopyright
\usepackage{booktabs}       
\usepackage{amsfonts}       
\usepackage{nicefrac}       
\usepackage{microtype}      
\usepackage{xcolor}         
\usepackage{tabularx}
\usepackage{multirow} 
\usepackage{makecell} 
\usepackage{amsmath}
\usepackage{xspace}
\usepackage{algorithm}

\usepackage[noend]{algpseudocode}

\newcommand{\ie}{i.e.\xspace}
\newcommand{\eg}{e.g.\xspace}

\title{ISO-RAG: Isoperimetric Noise Control for Retrieval-Augmented Generation}

\author{
    Siyuan Zhang\textsuperscript{\rm 1}, 
    Hanchen Wang\textsuperscript{\rm 1}, 
    Dong Wen\textsuperscript{\rm 2}, 
    Ying Zhang\textsuperscript{\rm 1}, 
    Wenjie Zhang\textsuperscript{\rm 2}
}
\affiliations{
    \textsuperscript{\rm 1}University of Technology Sydney \quad
    \textsuperscript{\rm 2}University of New South Wales
    
}

\begin{document}

\maketitle

\begin{abstract}
Retrieval-Augmented Generation (RAG) mitigates large language models (LLMs) hallucinations, yet conventional dense retrieval struggles with the complex reasoning paths of multi-hop question answering (QA). Graph-based RAG captures multi-step relationships but suffers from severe semantic drift and high online latency due to noisy global graph traversals. Thus, we propose \textbf{ISO-RAG} (\textbf{ISO}perimetric \textbf{R}etrieval-\textbf{A}ugmented \textbf{G}eneration), a training-free, purely topology-driven RAG framework. Breaking away from computationally expensive continuous geometric embeddings, ISO-RAG leverages discrete graph theory, specifically the local Cheeger ratio (topological expansion rate), to compute node-wise isoperimetric profiles. By identifying and pruning spurious shortcut edges that lead to combinatorial explosion, ISO-RAG restricts the search space to a strictly localized, contextually safe subgraph. This topological purification regulates deterministic Personalized PageRank (PPR) diffusion during retrieval, ensuring exact and low-latency convergence without probability leakage. Experiments on multi-hop QA benchmarks demonstrate that ISO-RAG outperforms state-of-the-art baselines by average absolute gains of 10.0\% in retrieval recall and 4.3\% in downstream exact match, achieving a superior accuracy-efficiency trade-off by fundamentally eliminating the latency bottleneck of global traversals. Our source code is available at \url{https://github.com/ZaiizaiZHANG/ISO-RAG.git}.
\end{abstract}

\section{Introduction}
\label{sec:introduction}

\begin{figure*}[t] 
  \centering
  \includegraphics[width=0.7\textwidth]{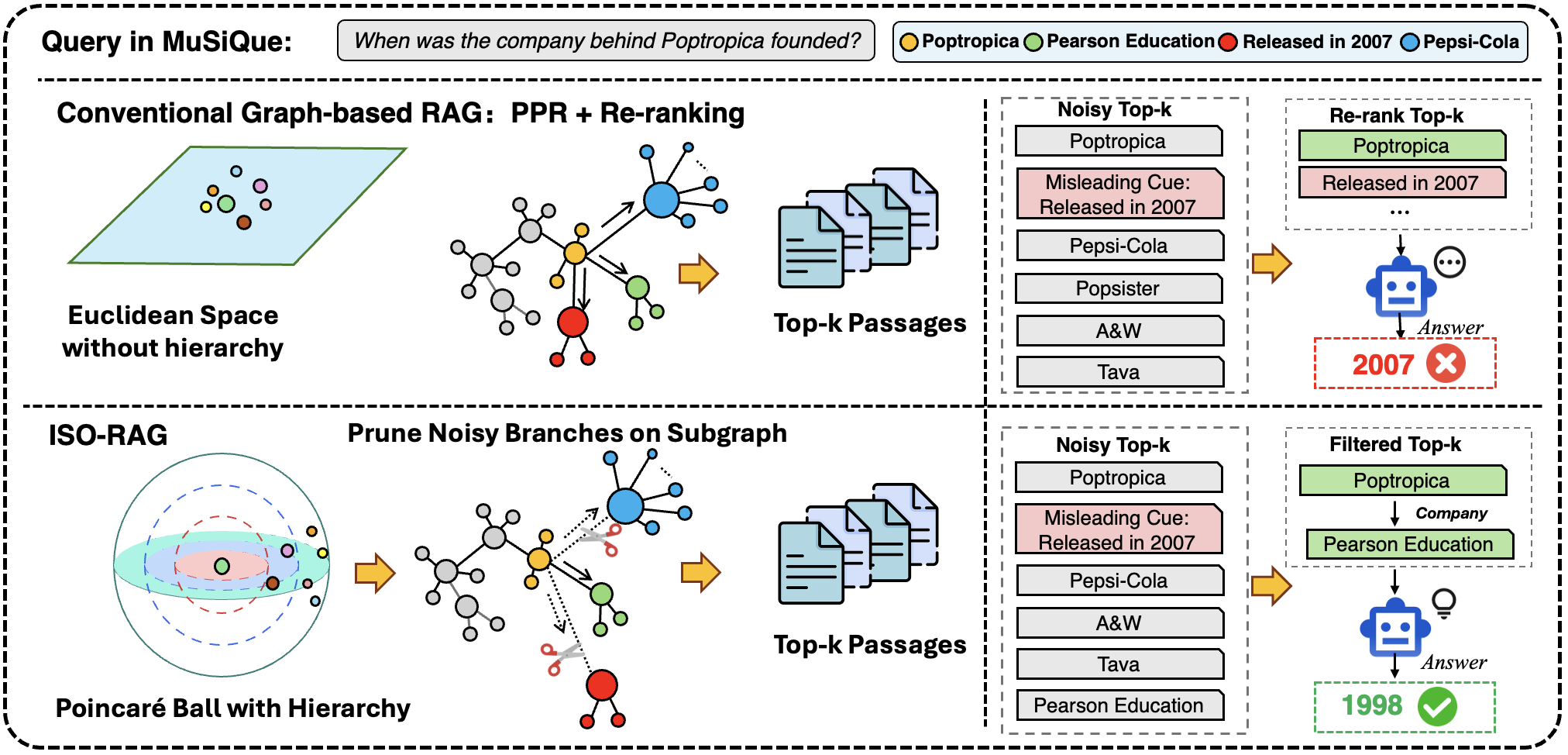} 
  \caption{{Conceptual comparison of RAG paradigms.} Top: Conventional graph-based RAG suffers from semantic drift: unconstrained diffusion allows retrieval signals to leak into irrelevant noise (blue nodes) and misleading cues (red nodes). Bottom: ISO-RAG leverages hyperbolic hierarchy and an isoperimetric control mechanism (scissors) to explicitly sever these erroneous branches. This topological intervention isolates a pure reasoning path (green nodes).}
  \label{fig:Introduction}
\end{figure*}

Retrieval-Augmented Generation (RAG)~\cite{lewis2020retrieval, guu2020realm} is a standard paradigm for improving the factuality and controllability of large language models (LLMs)~\cite{kaplan2020scaling,vaswani2017attention}. Grounding generation in external evidence substantially reduces hallucinations~\cite{Ji_2023, gao2023retrieval, jiang2023active} and improves answer reliability. While effective for single-hop factual lookup, RAG remains less reliable for multi-hop question answering, which requires connecting multiple evidence pieces scattered across different documents. In such settings, retrieval quality becomes the dominant bottleneck, as LLMs cannot reliably infer answers~\cite{wei2022chain, yao2022react, yao2023tree} without the complete reasoning chain.

Existing sparse and dense retrievers (e.g., BM25~\cite{robertson2009probabilistic}, MDR~\cite{xiong2020answering}, \cite{karpukhin2020dense}) rely on flat similarity matching, often failing to reconstruct the full reasoning chains required by compositional benchmarks~\cite{trivedi2022musique, chen2023benchmarking}. To address this, recent graph-based~\cite{edge2024graphrag, gutierrez2024hipporag, cao2025hyperbolicrag} methods organize corpora into interconnected structures for multi-hop evidence aggregation. Notably, HyperbolicRAG~\cite{cao2025hyperbolicrag} embeds document networks into continuous hyperbolic spaces to model their inherent scale-free hierarchies. This non-Euclidean mapping enables capturing complex multi-hop dependencies with low structural distortion.

Despite these advantages, conventional graph paradigms lack explicit topological intervention, manifesting in three limitations (Figure~\ref{fig:Introduction}): 
(1) Unconstrained diffusion leads to semantic drift. Graph-based frameworks~\cite{edge2024graphrag, gutierrez2024hipporag, cao2025hyperbolicrag} utilizing Personalized PageRank (PPR)~\cite{page1999pagerank} propagate probabilities over dense graphs. As illustrated by the MuSiQue query, answering requires traversing a specific two-hop path to the target entity (\eg, \textit{Poptropica} to \textit{Pearson Education}). However, unconstrained PPR retrieves misleading cues through spurious edges (\eg, release year \textit{2007}). Because these distractors exhibit high textual overlap with the query context, they bypass downstream re-rankers, causing the LLM to hallucinate the incorrect year. Crucially, if unconstrained diffusion breaks down on a mere two-hop path, this semantic drift compounds for more complex three- or four-hop queries.
(2) Continuous models fail to prune noise. Methods like HyperbolicRAG~\cite{cao2025hyperbolicrag} rely on continuous node embeddings without explicitly pruning noisy edges (depicted as the scissors in Figure~\ref{fig:Introduction}), thereby retaining erroneous pathways to distractors. This fails to isolate the specific reasoning branches necessary for accurate multi-hop deduction. 
(3) Dense graphs degrade efficiency. Computing random walks over the entire graph explores unrelated entities (\eg, distant brands like \textit{Pepsi-Cola}), which incurs significant computational overhead and dilutes the probability mass of the actual target entity. 
Consequently, these paradigms struggle to balance signal fidelity and retrieval latency.

In response to these limitations, we propose \textbf{ISO-RAG} (\textbf{ISO}perimetric \textbf{R}etrieval-\textbf{A}ugmented \textbf{G}eneration), a geometry-aware local graph retrieval framework for multi-hop QA. 
The core intuition of ISO-RAG is transitioning from unconstrained probability diffusion to geometrically regulated local diffusion. After routing a query to semantically aligned seed nodes to form a candidate subgraph, the framework maps these nodes into a hyperbolic space using the Poincar\'{e} ball model~\cite{nickel2017poincare, chami2019hyperbolic, balazevic2019multi, gulcehre2019hyperbolic, nickel2018learning, peng2022hyperbolic}. 
Document networks in multi-hop QA naturally form hierarchical structures, where a few generic entities act as dense hubs connecting numerous specific facts. Because hyperbolic space expands exponentially, it embeds these scale-free topologies with low geometric distortion. Crucially, this non-Euclidean embedding structurally highlights intrinsic bottlenecks~\cite{alon2021on, topping2021understanding}, such as the dense generic hubs~\cite{girvan2002community} that misguide retrieval. In spectral graph theory, the classical Cheeger constant~\cite{chung1997spectral} identifies global graph bottlenecks. Building on this geometric foundation, ISO-RAG introduces a localized isoperimetric control mechanism~\cite{andersen2006local,krioukov2010hyperbolic}.Because computing a strict set-level isoperimetric constant is computationally prohibitive for dynamic online retrieval, we design a numerically stable, node-wise proxy. This proxy acts as a structural filter, explicitly identifying and severing incompatible edges prior to propagation. By anchoring the Personalized PageRank diffusion to initial query-aligned seed passages and executing it strictly within this geometrically bounded subgraph, the framework mitigates semantic drift and facilitates noise-controlled evidence aggregation.

Our contributions are summarized as follows. 
\textbf{(1)} We propose \textbf{ISO-RAG}, a novel geometry-aware RAG framework that uses explicit topological control to mitigate spurious diffusion.
\textbf{(2)} At the core of this framework, we introduce an \textbf{isoperimetric control mechanism} that explicitly prunes misleading connections to dense hubs prior to diffusion.
\textbf{(3)} Extensive evaluations demonstrate that ISO-RAG achieves a highly favorable balance between \textbf{retrieval efficiency} and \textbf{downstream QA performance}, delivering robust average absolute gains of nearly 10\% in Recall@5 and 4.3\% in Exact Match over competitive baselines.

\section{Related Works}

\textbf{Sparse and Dense Retrieval for Multi-Hop QA.}
Conventional retrieval encompasses sparse methods like BM25 \cite{robertson2009probabilistic} and dense models ranging from flat bi-encoders \cite{karpukhin2020dense} to multi-hop extensions like MDR \cite{xiong2020answering}. Whether utilizing exact keyword matching or cosine similarity, these approaches operate in fundamentally flat search spaces. Compressing documents into isolated points optimized for direct semantic overlap, dense vectors cannot explicitly model relationships between intermediate entities. This structural limitation fragments reasoning by allowing lexically similar yet logically disconnected distractors to overshadow critical evidence. Consequently, flat search spaces struggle with the compositional reasoning paths required by increasingly difficult multi-hop QA datasets like HotpotQA \cite{yang2018hotpotqa}, 2WikiMultihopQA \cite{ho2020constructing}, and MuSiQue \cite{trivedi2022musique}.

\textbf{Conventional Graph-based RAG Systems.}
To overcome flat semantic spaces, graph-based retrieval structures corpora into networks capturing multi-hop dependencies. Notable architectures include GraphRAG \cite{edge2024graphrag} utilizing hierarchical summaries with optional heuristic routing, LightRAG \cite{guo2024lightrag} employing dual-level structures, and HippoRAG2 \cite{gutierrez2025from} leveraging neurobiologically-inspired memory networks with continuous activation spreading. Despite improving recall, these baselines rely on heuristic edge weighting and unconstrained probability diffusion; for instance, HippoRAG2 executes unrestricted global PPR. Consequently, this uncontrolled diffusion introduces severe topological noise during retrieval \cite{alon2021on, topping2021understanding}. Without rigorous mathematical bounds pruning the search space, these frameworks inevitably retrieve spurious subgraphs and misleading entities before generation.

\textbf{Hyperbolic Geometry and Continuous Aggregation.}
The inherent hierarchical structure of knowledge graphs makes them poorly suited for Euclidean embeddings \cite{nickel2017poincare, sala2018representation}. Foundational graph neural networks therefore extend representation learning into hyperbolic space through architectures like HGCN \cite{chami2019hyperbolic} and related hyperbolic networks \cite{gulcehre2019hyperbolic, peng2022hyperbolic}. Mainstream methodologies use the Poincar'{e} ball model \cite{nickel2017poincare} for its intuitive conformal geometry, or the Lorentz model \cite{nickel2018learning} for its numerical advantages in distance optimization. Both models provide exponential capacity to embed complex networks with minimal structural distortion. Recent works like HyperbolicRAG \cite{cao2025hyperbolicrag} attempt to leverage this by introducing hyperbolic representations into RAG. However, despite operating in hyperbolic space, its probability diffusion remains fundamentally unconstrained. By performing continuous neighborhood aggregation without explicit discrete pruning, HyperbolicRAG fails to resolve the topological bottleneck, inevitably accumulating topological noise and causing severe semantic drift during retrieval.

\section{Methodology}
\label{sec:methodology}

We present \textbf{ISO-RAG} (\textbf{ISO}perimetric \textbf{R}etrieval-\textbf{A}ugmented \textbf{G}eneration), a geometry-aware local graph retrieval framework for multi-hop question answering. The core idea is to avoid broad graph-wide diffusion by combining four components: (i) query-aware seed routing, (ii) local candidate graph construction, (iii) isoperimetric structural filtering derived from hyperbolic structure, and (iv) deterministic personalized PageRank on the filtered local graph.

\subsection{Problem Formulation}
\label{subsec:task_setup}

Let $\mathcal{C} = \{p_1, p_2, \dots, p_N\}$ denote a corpus of passages. Given a multi-hop query $q$, the retrieval objective is to extract a top-$K$ subset $\mathcal{R}_K(q) \subset \mathcal{C}$, such that the retrieved passages jointly cover the evidence required to answer $q$.

We model the corpus as an undirected retrieval graph $\mathcal{G} = (\mathcal{V}, \mathcal{E})$, where each node $u \in \mathcal{V}$ represents a passage equipped with a dense semantic embedding $\mathbf{h}_u \in \mathbb{R}^d$. $\mathcal{G}$ serves as a question-induced passage co-occurrence graph: an edge $(u,v) \in \mathcal{E}$ is established whenever passages $u$ and $v$ co-occur within a training instance or retrieval context.

Such co-occurrence graphs effectively expose latent multi-hop dependencies, but they also inherently introduce noisy shortcuts and hub-like regions. As a result, unconstrained diffusion processes can prematurely leak probability mass into generic yet weakly useful passages, reducing both retrieval precision and downstream QA performance.

\subsection{Seeded Local Graph Construction}
\label{subsec:overview}

Given a query $q$, ISO-RAG first retrieves its dense embedding $\mathbf{h}_q$ from a precomputed embedding cache and measures its cosine similarity to every node embedding:

\begin{equation}
s_u(q) = \mathrm{cos}(\mathbf{h}_q, \mathbf{h}_u), \qquad u \in \mathcal{V}.
\end{equation}

Let $\mathcal{S}(q)$ denote the set of top-$m$ seed nodes selected according to $s_u(q)$:
\begin{equation}
\mathcal{S}(q) = \operatorname*{top-m}_{u \in \mathcal{V}} \; s_u(q).
\end{equation}

Rather than propagating over the full graph, ISO-RAG restricts the search space to a compact local candidate set $\mathcal{V}_{\mathrm{loc}}(q) \subseteq \mathcal{V}$. To ensure both high recall and structural connectivity, we construct this set by integrating two components: a dense semantic pool and a topology-aware neighborhood. 

Specifically, let $\mathcal{V}_{\mathrm{dense}}(q)$ denote the top-$L$ passages retrieved via $s_u(q)$ (where $L \ge m$, thus containing the seed set $\mathcal{S}(q)$). Let $\mathcal{V}_{\mathrm{expand}}(q)$ denote the $k$-hop structural neighborhood expanded from $\mathcal{S}(q)$. The local candidate set is defined as their union:
\begin{equation}
    \mathcal{V}_{\mathrm{loc}}(q) = \mathcal{V}_{\mathrm{dense}}(q) \cup \mathcal{V}_{\mathrm{expand}}(q).
    \label{eq:local_candidate_set}
\end{equation}
The induced local subgraph is $\mathcal{G}_{\mathrm{loc}}(q) = \mathcal{G}[\mathcal{V}_{\mathrm{loc}}(q)]$.

This query-conditioned local graph substantially reduces the search space and transforms retrieval from global diffusion vulnerable to topological noise into local propagation anchored at semantically aligned entry points.

\subsection{Hyperbolic Isoperimetric Edge Filtering}
\label{subsec:isoperimetric_filtering}

\textbf{Hyperbolic structural signal.} To characterize local graph structure, we map node representations into the Poincar\'e ball~\cite{nickel2017poincare,chami2019hyperbolic,balazevic2019multi}:
\begin{equation}
\mathbf{z}_u = f_\theta(\mathbf{h}_u), \qquad \mathbf{z}_u \in \mathbb{B}^d,
\end{equation}
where
\begin{equation}
\mathbb{B}^d = \left\{ \mathbf{z}\in\mathbb{R}^d : \|\mathbf{z}\|_2 < 1 \right\},
\end{equation}
and $\|\cdot\|_2$ denotes the standard Euclidean norm. The mapping $f_\theta$, which includes a projection onto the open unit ball, directly maps precomputed Euclidean text embeddings into the hyperbolic manifold. To align textual semantics with discrete multi-hop topology, $f_\theta$ is trained offline via a graph-supervised margin-based triplet loss and a radial depth regularizer (detailed in Appendix). This helps ensure the representations encapsulate both raw semantics and hierarchical structures.

For a node $u$, let
\begin{equation}
\lambda(\mathbf{z}_u) = \frac{2}{1-\|\mathbf{z}_u\|_2^2}
\end{equation}
denote the conformal factor of the Poincar\'e metric at $\mathbf{z}_u$. In Riemannian geometry, this factor dictates the volume expansion of the local space. Therefore, the geometric volume occupied by a node is intrinsically driven by this conformal scaling. We thus define the local volume proxy as:
\begin{equation}
\mathrm{v}(u) = \lambda(\mathbf{z}_u)^p.
\end{equation}
Geometrically, $\mathrm{v}(u)$ quantifies this continuous spatial occupancy, which acts as an inverse indicator of semantic breadth. Due to the exponential outward expansion of the Poincar\'e ball, generic semantic hubs at the origin are tightly compressed into minimal conformal volumes, whereas highly specific factual entities at the periphery occupy vast spatial regions. In strict Riemannian geometry, the true volume scales with the dimensionality $d$. However, computing $\lambda(\mathbf{z}_u)^d$ for high-dimensional embeddings ($d \ge 128$) inevitably leads to numerical overflow. To ensure system reliability, we introduce a tunable scaling exponent $p \ll d$. This engineering formulation provides a numerically stable proxy for the node volume that flexibly controls the dynamic range of the structural signal. By doing so, we preserve the core monotonic conformal scaling property of hyperbolic space while guaranteeing robust online computation.

To construct a principled structural ratio, we define a localized geometric measure based on the classical Cheeger constant. For any node $u \in \mathcal{V}$, let $S_u = \{u\} \cup \mathcal{N}(u)$ denote its closed 1-hop neighborhood set. We define the internal geometric volume of this local region as the sum of the conformal volume proxies of its constituent nodes:
\begin{equation}
V(S_u) = \sum_{n \in S_u} \mathrm{v}(n).
\end{equation}

Next, we identify the topological boundary of this region. Let $\partial S_u$ denote the 2-hop boundary shell of $u$, consisting of all nodes adjacent to $\mathcal{N}(u)$ that are not contained within $S_u$. The geometric volume of this boundary is analogously defined as:
\begin{equation}
V(\partial S_u) = \sum_{n \in \partial S_u} \mathrm{v}(n).
\end{equation}

Having formalized both the internal and boundary volumes using the exact same hyperbolic measure, we define the node-wise structural pruning score $\phi_u$ as the localized geometric Cheeger ratio:
\begin{equation}
\phi_u = \frac{V(\partial S_u)}{V(S_u) + \varepsilon},
\end{equation}
where $\varepsilon > 0$ is a small stability constant. Unlike heuristic formulations that mix mismatched topological and geometric scales, this definition strictly preserves the isoperimetric nature of the score. The proxy measures the relative geometric expansion of a node's local neighborhood. Generic hubs exhibit explosive boundary volumes ($V(\partial S_u)$) compared to their highly compressed internal neighborhood volumes ($V(S_u)$), yielding extremely large $\phi_u$ values. By evaluating this rigorous volume-to-volume ratio, the framework can explicitly identify structural bottlenecks without requiring computationally prohibitive global graph partitioning.

\textbf{Discriminative Power of the Proxy.} In scale-free hyperbolic embeddings, a node's radial distance inversely tracks its topological degree, while its conformal volume grows exponentially with radius. This dual scaling creates a geometric mismatch for generic hubs: they reside near the origin with heavily compressed individual volumes, yet their massive topological connectivity bridges to numerous specific nodes at the periphery. Consequently, for a central hub, its 2-hop boundary shell $\partial S_u$ reaches into the expansive periphery, accumulating an explosive boundary volume $V(\partial S_u)$, while its internal 1-hop neighborhood volume $V(S_u)$ remains heavily constrained. This extreme volume-to-volume divergence triggers massive $\phi_u$ anomalies for spurious edges connecting factual nodes to unrelated hubs, enabling ISO-RAG to structurally isolate probability leakage. Detailed mathematical formulations are provided in Appendix.

\textbf{Edge Compatibility Filtering.} To prune spurious edges bridging structurally dissimilar nodes, we enforce structural consistency between endpoints. An edge $(u,v) \in \mathcal{G}_{\mathrm{loc}}(q)$ is retained only if:
\begin{equation}
\label{eq:filtering}
\frac{\min(\phi_u,\phi_v)}{\max(\phi_u,\phi_v)} \ge \beta,
\end{equation}
where $\beta \in (0,1]$ is a validation-tuned tolerance threshold. Equivalently, in log-scale:
\begin{equation}
|\log \phi_u - \log \phi_v| \le -\log \beta.
\end{equation}
Here, $-\log \beta$ bounds the maximum allowable structural divergence. Valid reasoning steps between nodes of comparable specificity exhibit small $\phi$-divergences. Conversely, edges bridging opposite structural extremes (e.g., direct transitions between specific factual leaves and generic hubs) inevitably violate this threshold and are systematically pruned. This dual-sided filtering mitigates semantic drift from two directions: it blocks forward probability leakage into hubs during propagation, and isolates erroneously retrieved hub anchors before diffusion begins. Because all node-wise structural scores $\phi_u$ can be fully precomputed and cached, this geometric filtering introduces negligible online computational latency. We provide details in Appendix.

Let
\begin{equation}
\widetilde{\mathcal{G}}_{\mathrm{loc}}(q)
=
\left(
\mathcal{V}_{\mathrm{loc}}(q),
\widetilde{\mathcal{E}}_{\mathrm{loc}}(q)
\right)
\end{equation}
denote the resulting geometrically filtered local graph, which provides a structurally coherent and bounded manifold for the subsequent localized PageRank.

\subsection{Localized Deterministic Personalized PageRank}
\label{subsec:localized_ppr}

After edge filtering, retrieval operates strictly on the compact graph $\widetilde{\mathcal{G}}_{\mathrm{loc}}(q)$. We re-normalize its adjacency matrix to derive a valid column-stochastic transition matrix $\widetilde{\mathbf{W}}$. 

The personalization vector $\mathbf{r}(q)$ distributes the initial probability mass exclusively across the selected seed nodes $\mathcal{S}(q)$ via a temperature-scaled softmax:
\begin{equation}
\label{eq:seed_softmax}
r_u(q)=
\begin{cases}
\displaystyle
\frac{\exp(\tau \cdot s_u^{\text{dense}})}
{\sum_{v\in\mathcal{S}(q)} \exp(\tau \cdot s_v^{\text{dense}})}, & u\in\mathcal{S}(q), \\[1.2em]
0, & \text{otherwise},
\end{cases}
\end{equation} 
where $\tau > 0$ controls the mass concentration sharpness, and $s_u^{\text{dense}}$ denotes the initial dense semantic similarity score.

The localized PPR vector $\boldsymbol{\pi}(q)$ is defined as the unique fixed point of the diffusion process:
\begin{equation}
\boldsymbol{\pi}(q)
=
(1-\alpha)\mathbf{r}(q)
+
\alpha \widetilde{\mathbf{W}} \boldsymbol{\pi}(q),
\end{equation}
where $\alpha \in (0,1)$ is the damping factor (with $1-\alpha$ acting as the teleportation probability). During diffusion, the probability mass of dangling nodes is intrinsically redistributed via $\mathbf{r}(q)$, which falls back to a uniform distribution if seed weights degrade to zero. Rather than relying on stochastic random walks that introduce approximation variance, we compute $\boldsymbol{\pi}(q)$ deterministically via power iteration. Because our structural filtering strictly bounds the diffusion space to the compact local subgraph $\widetilde{\mathcal{G}}_{\mathrm{loc}}(q)$, this exact computation remains highly efficient and yields stable structural scores.

Finally, the converged structural scores are fused with the initial dense semantic similarities. Denoting the scalar structural score for node $u$ as $\pi_u(q)$, which is extracted from the vector $\boldsymbol{\pi}(q)$, the final retrieval score $s_u^{\text{final}}$ for each passage $u$ is computed via a linear combination:
\begin{equation}
\label{eq:fusion}
s_u^{\text{final}} = \lambda_{\mathrm{g}} \pi_u(q) + \lambda_{\mathrm{d}} s_u^{\text{dense}},
\end{equation}
where $\pi_u(q)$ and $s_u^{\text{dense}}$ are min-max normalized over the candidate set prior to fusion, and $\lambda_{\mathrm{g}}$ and $\lambda_{\mathrm{d}}$ are tunable balancing weights. The candidate passages are subsequently ranked by $s_u^{\text{final}}$ to extract the optimal top-$K$ evidence set $\mathcal{R}_K(q)$ for downstream QA. This dual-signal fusion elegantly couples the semantic recall of dense models with the structural multi-hop precision of our isoperimetric framework, ensuring that the final ranking is both contextually relevant and topologically coherent.

\section{Experiments}
\begin{table}[!t]
\centering
\small
\setlength{\tabcolsep}{1.2pt}
\begin{tabular}{ll cccccc}
\toprule
\multirow{2.5}{*}{\textbf{Model}} & \multirow{2.5}{*}{\textbf{Retriever}} & \multicolumn{2}{c}{\textbf{HotpotQA}} & \multicolumn{2}{c}{\makecell{\textbf{2Wiki-}\\\textbf{MultihopQA}}} & \multicolumn{2}{c}{\textbf{MuSiQue}} \\
\cmidrule(lr){3-4} \cmidrule(lr){5-6} \cmidrule(lr){7-8}
& & F1 & EM & F1 & EM & F1 & EM \\
\midrule

\multirow{6}{*}{Qwen2.5} 
& GraphRAG      & 79.1 & 72.3 & 54.1 & 52.8 & 28.8 & 23.2 \\
& GraphRAG+PPR  & 75.0 & 68.4 & 54.5 & 52.9 & 30.6 & 24.2 \\
& LightRAG      & 79.8 & 72.3 & 72.2 & 68.9 & 34.0 & 28.4 \\
& HippoRAG2     & 80.6 & 73.1 & 62.3 & 59.7 & 36.9 & 30.8 \\
& HyperbolicRAG & 79.6 & 72.3 & 60.0 & 58.2 & 30.6 & 25.6 \\
& \textbf{ISO-RAG} & \textbf{81.1} & \textbf{74.1} & \textbf{76.9} & \textbf{73.2} & \textbf{40.1} & \textbf{33.9} \\
\midrule

\multirow{6}{*}{\begin{tabular}{@{}l@{}}Qwen-\\ Plus\end{tabular}} 
& GraphRAG      & 80.4 & 72.8 & 56.7 & 55.2 & 28.7 & 23.3 \\
& GraphRAG+PPR  & 78.1 & 70.8 & 57.5 & 55.2 & 30.1 & 24.0 \\
& LightRAG      & 82.3 & \textbf{75.0} & 72.8 & 68.8 & 30.9 & 24.6 \\
& HippoRAG2     & 82.2 & 74.6 & 63.7 & 60.9 & 34.2 & 27.7 \\
& HyperbolicRAG & 81.4 & 73.7 & 61.6 & 59.4 & 32.1 & 25.4 \\
& \textbf{ISO-RAG} & \textbf{82.5} & \textbf{75.0} & \textbf{80.4} & \textbf{75.8} & \textbf{42.8} & \textbf{31.4} \\
\midrule

\multirow{6}{*}{\begin{tabular}{@{}l@{}}Qwen3-\\ Max\end{tabular}} 
& GraphRAG      & 83.1 & 76.0 & 60.9 & 59.2 & 28.7 & 23.3 \\
& GraphRAG+PPR  & 81.0 & 74.1 & 59.5 & 57.7 & 30.1 & 24.0 \\
& LightRAG      & 84.6 & 77.4 & 77.0 & 73.8 & 33.6 & 28.3 \\
& HippoRAG2     & 84.8 & 77.7 & 67.4 & 64.6 & 36.9 & 30.9 \\
& HyperbolicRAG & 83.9 & 76.5 & 64.5 & 62.5 & 34.6 & 28.8 \\
& \textbf{ISO-RAG} & \textbf{85.2} & \textbf{78.2} & \textbf{84.7} & \textbf{80.7} & \textbf{40.2} & \textbf{33.7} \\

\bottomrule
\end{tabular}
\caption{Main Results on Multi-Hop QA Datasets. F1 and EM scores are in percentage (\%). Best results are \textbf{bolded}.}
\label{tab:main_results}
\end{table}
\label{sec:experiments}


In this section, we comprehensively evaluate ISO-RAG to answer the following Research Questions (RQs): \textbf{RQ1 (Overall Performance):} Does ISO-RAG outperform existing dense and graph-based retrieval methods in both retrieval accuracy and downstream multi-hop QA? \textbf{RQ2 (Efficiency):} Can the localized deterministic routing paradigm achieve better retrieval-time efficiency compared to unconstrained graph traversals? \textbf{RQ3 (Geometric Filtering):} How does the isoperimetric signal ($\phi$) explicitly contribute to noise-aware structural filtering?

\begin{table*}[!htbp] 
\centering
\small

\begin{tabular}{l cccc cccc cccc}
\toprule
\multirow{2.5}{*}{\textbf{Method}} & \multicolumn{4}{c}{\textbf{HotpotQA}} & \multicolumn{4}{c}{\textbf{2WikiMultihopQA}} & \multicolumn{4}{c}{\textbf{MuSiQue}} \\
\cmidrule(lr){2-5} \cmidrule(lr){6-9} \cmidrule(lr){10-13}
& R@5 & R@10 & P@5 & P@10 & R@5 & R@10 & P@5 & P@10 & R@5 & R@10 & P@5 & P@10 \\
\midrule

BM25               & 62.70 & 75.25 & 25.08 & 15.05 & 39.80 & 48.20 & 18.42 & 11.08 & 27.00 & 32.15 & 10.64 & 6.34 \\
Flat Dense         & 81.90 & 87.95 & 32.76 & 17.59 & 69.43 & 71.58 & 31.82 & 16.39 & 52.40 & 58.95 & 20.54 & 11.57 \\
MDR                & 78.80 & 88.15 & 31.52 & 17.63 & 61.65 & 68.55 & 27.70 & 15.53 & 51.70 & 59.45 & 20.34 & 11.70 \\
\midrule

Vanilla PPR        & 81.75 & 94.40 & 32.70 & 18.88 & 69.05 & 78.95 & 31.80 & 18.96 & 70.10 & 77.30 & 27.12 & 14.98 \\
GraphRAG           & 85.25 & 96.20 & 34.10 & 19.24 & 71.15 & 88.45 & 32.84 & 21.34 & 59.00 & 72.30 & 22.84 & 13.99 \\
GraphRAG+PPR       & 71.45 & 94.20 & 28.58 & 18.84 & 69.78 & 81.95 & 32.38 & 19.91 & 63.20 & 76.15 & 24.40 & 14.74 \\
LightRAG           & 87.30 & 97.20 & 34.92 & 19.44 & 77.93 & 88.80 & 35.92 & 20.95 & 55.05 & 67.05 & 21.60 & 13.16 \\
HippoRAG2          & 86.65 & 97.85 & 34.66 & 19.57 & 70.18 & 82.80 & 32.30 & 19.88 & 63.85 & 75.40 & 26.14 & 14.64 \\
HyperbolicRAG      & 85.30 & 96.85 & 34.12 & 19.37 & 70.45 & 78.38 & 32.34 & 18.55 & 55.35 & 76.05 & 21.70 & 14.70 \\
\midrule

\textbf{ISO-RAG}   & \textbf{88.05} & \textbf{98.05} & \textbf{35.22} & \textbf{19.61} & \textbf{88.10} & \textbf{97.15} & \textbf{40.10} & \textbf{22.90} & \textbf{74.75} & \textbf{84.00} & \textbf{28.98} & \textbf{16.31} \\

\bottomrule
\end{tabular}
\caption{Retrieval performance comparison of Recall (R@) and Precision (P@) at top-5 and top-10 across three multi-hop datasets. Best results are \textbf{bolded}.}
\label{tab:retrieval_main}
\end{table*}

\subsection{Experimental Setup}

\textbf{Datasets \& Graph Construction.} We evaluate ISO-RAG on three standard multi-hop QA benchmarks with varying reasoning complexities: HotpotQA~\cite{yang2018hotpotqa} (primarily 2-hop), 2WikiMultihopQA~\cite{ho2020constructing} (2--4 hop), and MuSiQue~\cite{trivedi2022musique} (up to 4-hop compositional reasoning). To establish a unified evaluation setting for both structural retrieval and end-to-end QA, we randomly sample 1,000 instances from the validation set of each benchmark. For graph construction, rather than building a traditional entity-relation knowledge graph, we construct a question-induced passage co-occurrence graph (details in Appendix. Passages are treated as nodes, with edges connecting passages that co-occur in the same training instance. Node texts are embedded offline using the \texttt{text-embedding-v3} encoder~\cite{zhang2025qwen3}. To strictly prevent data leakage, the passage co-occurrence graphs are constructed exclusively using the training splits of the respective datasets. Validation and test sets are completely excluded from the graph construction phase, ensuring that the retrieval framework does not benefit from any benchmark-specific structural shortcuts.

Table~\ref{tab:efficiency_all_datasets} compares retrieval latency and token usage under Qwen3-Max to assess the computational advantage of our localized pipeline.

\textbf{Baselines.} To comprehensively evaluate retrieval quality, we categorize baselines into non-graph methods (BM25~\cite{robertson2009probabilistic}, Flat Dense~\cite{karpukhin2020dense}, MDR~\cite{xiong2020answering}) and graph-based paradigms. The latter ranges from foundational algorithms (Vanilla PPR~\cite{page1999pagerank}) to state-of-the-art frameworks (GraphRAG~ \cite{edge2024graphrag}, GraphRAG+PPR~\cite{page1999pagerank}, LightRAG~\cite{guo2024lightrag}, HippoRAG2~\cite{gutierrez2024hipporag}, and HyperbolicRAG~\cite{cao2025hyperbolicrag}). 
As Vanilla PPR is a foundational retrieval algorithm rather than an end-to-end RAG pipeline, we focus our QA assessment exclusively on the aforementioned state-of-the-art frameworks designed for generative tasks. To execute this generation, the retrieval outputs are paired with multiple LLM backbones (Qwen2.5, Qwen-Plus~\cite{qwen2024qwen25}, and Qwen3-Max~\cite{qwen2025qwen3}).

\textbf{Implementation Details.} We evaluate downstream QA performance using standard Exact Match (EM) and F1 scores. For fair comparison, all graph frameworks supply the raw text of their retrieved top-$k$ passages to the LLM generator using an identical prompt template. Comprehensive implementation details, including hyperparameter tuning (\eg, PageRank $\alpha$), exact grid-search ranges, and full prompt templates, are detailed in Appendix.

\subsection{Main Results: Retrieval and QA Performance (RQ1)}

We first evaluate the fundamental retrieval capability (Tables~\ref{tab:retrieval_main}) and the end-to-end QA generation quality (Table~\ref{tab:main_results}). ISO-RAG consistently yields the best results across all settings.

\textbf{Consistent Gains in Shallow Reasoning.} While performance on HotpotQA is nearing saturation for high-capacity models, ISO-RAG still guarantees stable improvements, achieving 88.05 R@5 (+0.75 absolute points over the strongest baseline) and peaking at 85.2 F1 score with Qwen3-Max. 
Notably, the improvements are robust across model scales, suggesting that our retrieval framework itself fundamentally drives the observed gains.

\textbf{Superiority in Complex Topologies.} Table~\ref{tab:main_results} demonstrates that ISO-RAG outperforms all baselines on 2WikiMultihopQA. Notably, it surpasses the strongest baseline, LightRAG, achieving an R@5 of 88.10 with a 10.17-point absolute improvement. 
Compared to recent topology-driven frameworks (\ie, HippoRAG2 and HyperbolicRAG), this gap widens, yielding a relative R@5 gain exceeding 25\%. Furthermore, QA evaluation using Qwen3-Max yields an F1 score of 84.7 and an EM score of 80.7. These results confirm the efficacy of ISO-RAG in structured multi-hop scenarios, where preserving valid intermediate hops and suppressing spurious diffusion are critical.

\textbf{Robustness against High Noise.} The MuSiQue dataset presents a severe challenge of deep compositional reasoning amidst dense distractors. In this regime, while ISO-RAG achieves a state-of-the-art R@10 score of 84.00 (+11.4\% relative gain over HippoRAG2) alongside highly competitive precision of 16.31, its full potential materializes in the QA phase. Evaluated with Qwen-Plus, the framework peaks at an F1 score of 42.8, securing an approximate 25\% relative gain over HippoRAG2. This disproportionate amplification, transitioning from a steady retrieval improvement to a drastic QA leap, demonstrates that ISO-RAG does not merely accumulate disjoint relevant documents. Instead, driven by its high retrieval precision, it successfully isolates the precise, noise-free multi-hop evidence pathways required to cross the reasoning threshold of the LLM.

\begin{figure*}[t] 
\centering
\includegraphics[width=0.75\textwidth]  
{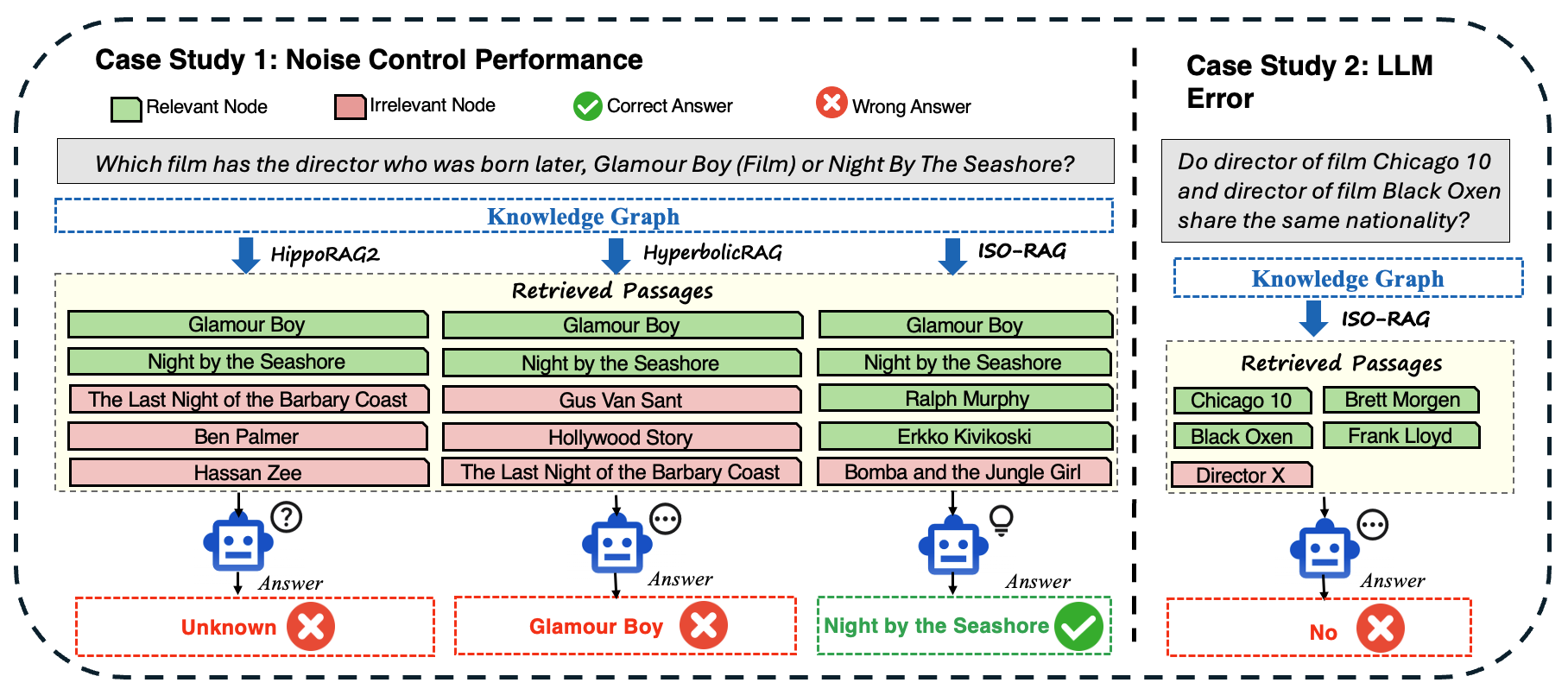}
\caption{Case Study}
\label{fig:phi_casestudy}
\end{figure*}

\subsection{Efficiency Analysis (RQ2)}

\begin{table}[t]
\centering
\small 
\setlength{\tabcolsep}{4pt} 
\renewcommand{\arraystretch}{0.9}
\begin{tabular}{llcc}
\toprule
\multirow{2.5}{*}{\textbf{Dataset}} & \multirow{2.5}{*}{\textbf{Method}} & \multicolumn{2}{c}{\textbf{Efficiency}} \\
\cmidrule(lr){3-4}
& & \textbf{Retr/Q (ms)} $\downarrow$ & \textbf{Avg Prompt} \\
\midrule
\multirow{6}{*}{HotpotQA} 
& GraphRAG & \textbf{10.8} & 1619.91 \\
& GraphRAG+PPR & 15.1 & 1652.10 \\
& LightRAG & 22.9 & 1644.83 \\
& HippoRAG2 & 52.0 & 895.16 \\
& HyperbolicRAG & 92.7 & \underline{889.61} \\
& \textbf{ISO-RAG} & \underline{12.3} & \textbf{886.44} \\
\midrule
\multirow{6}{*}{\begin{tabular}{@{}l@{}}2Wiki- \\ MultihopQA\end{tabular}} 
& GraphRAG & \textbf{6.2} & 1389.09 \\
& GraphRAG+PPR & \underline{8.7} & 1225.13 \\
& LightRAG & 14.4 & 1271.18 \\
& HippoRAG2 & 49.1 & \textbf{720.89} \\
& HyperbolicRAG & 71.1 & \underline{758.91} \\
& \textbf{ISO-RAG} & 11.5 & 793.49 \\
\midrule
\multirow{6}{*}{MuSiQue} 
& GraphRAG & \textbf{9.9} & 1579.64 \\
& GraphRAG+PPR & 16.2 & 1588.76 \\
& LightRAG & 65.6 & 1561.58 \\
& HippoRAG2 & 117.2 & \underline{912.46} \\
& HyperbolicRAG & 375.1 & \textbf{900.95} \\
& \textbf{ISO-RAG} & \underline{14.2} & 927.07 \\
\bottomrule
\end{tabular}
\caption{Efficiency Comparison across Datasets. Latency is measured in milliseconds per query (Retr/Q). Best results in each dataset are \textbf{bolded}, and second-best results are \underline{underlined}.}
\label{tab:efficiency_all_datasets}
\vspace{-1.8 em}
\end{table}
\begin{figure}[!hbtp] 
  \centering
  \includegraphics[width=\columnwidth]{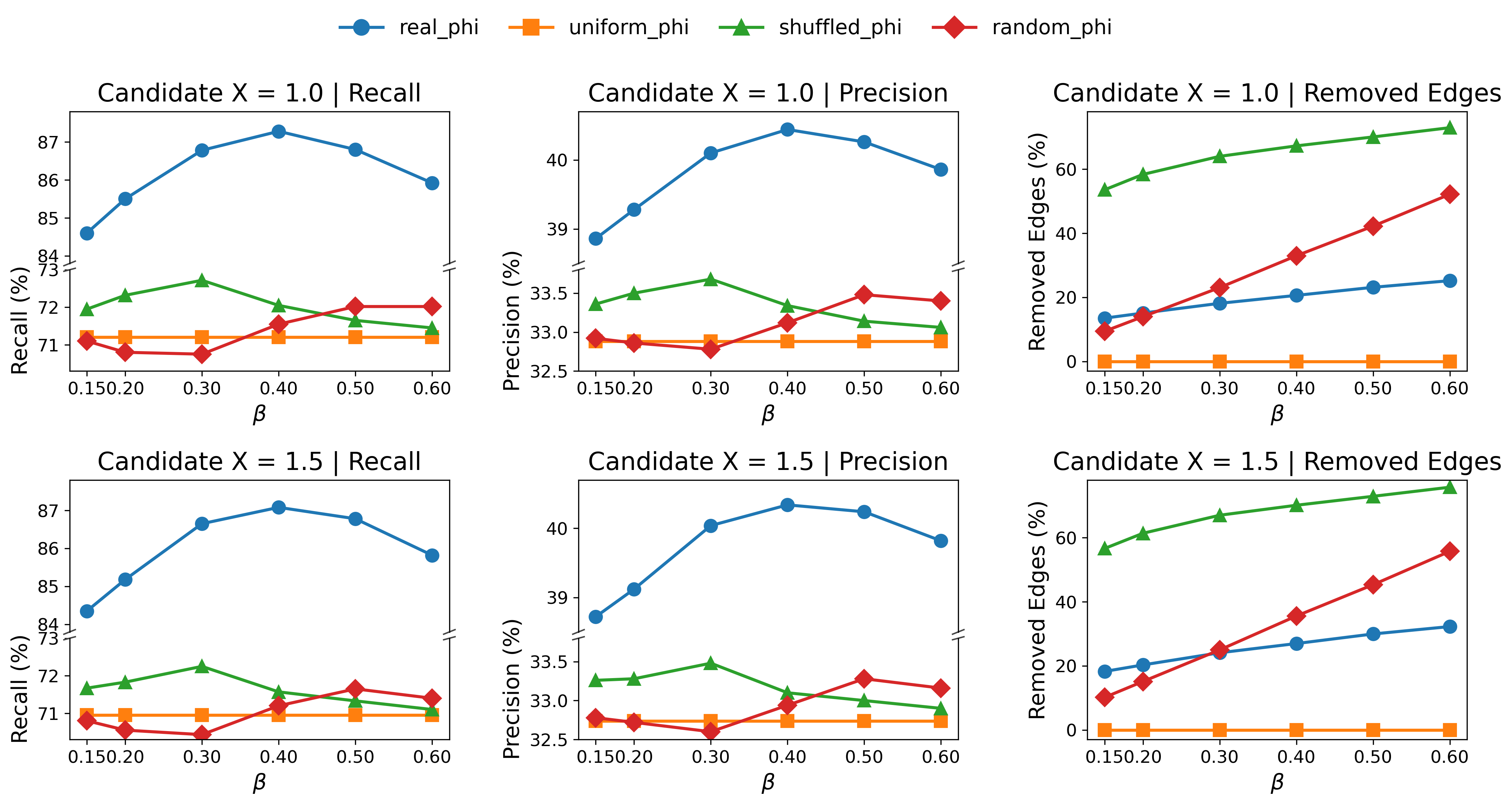}
  \caption{Retrieval Performance and Removed Edges Grouped by Floor ($\beta$) and Candidate $X$ Configurations (in \%)}
  \label{fig:phi_ablation}
\end{figure}

\textbf{Latency Reduction via Local Subgraphs.} ISO-RAG maintains stable millisecond-level speeds across all benchmarks, ranking as the second fastest retriever on both HotpotQA and MuSiQue. While marginally trailing the vanilla GraphRAG baseline in raw speed, it provides a substantially more precise reasoning context. Compared to recent topology-driven frameworks, the latency gap is particularly striking: on MuSiQue, ISO-RAG delivers an approximate 8$\times$ speedup over HippoRAG2 and operates over 25$\times$ faster than HyperbolicRAG. This confirms that bounding PageRank within a geometrically filtered subgraph successfully bypasses the heavy overhead of global traversals.

\textbf{Favorable Accuracy-Efficiency Trade-off.} ISO-RAG exhibits highly competitive token efficiency, incurring the lowest prompt token overhead among all baselines on HotpotQA. While structurally complex datasets require marginally more tokens than HippoRAG2 and HyperbolicRAG, this slight increment is fully offset by substantial gains in retrieval accuracy. Ultimately, this confirms that ISO-RAG delivers a strictly higher density of actionable reasoning chains per token, ensuring a highly cost-effective retrieval process.

\subsection{Ablation: Isoperimetric Geometric Filtering (RQ3)}

We isolate the isoperimetric filtering module on 2WikiMultihopQA, where the impact of geometric guidance is most pronounced, to compare true geometric guidance against unconstrained, random, or topologically decoupled edge pruning under highly noisy conditions (Figure~\ref{fig:phi_ablation}). 

We define three core variables to control the ablation space: Candidate $X$, the over-retrieval factor applied to the initial dense pool prior to $\phi$-filtering; $\beta$, the threshold controlling geometric edge pruning strictness; and four $\phi$ mapping strategies: (1) \texttt{real\_phi} applies actual learned scores to capture true local continuity; (2) \texttt{uniform\_phi} assigns a constant value, disabling the filter to revert to vanilla PPR; (3) \texttt{shuffled\_phi} randomly permutes \texttt{real\_phi} values, preserving global distribution but destroying correlation with graph topology; and (4) \texttt{random\_phi} applies uniform random values to test arbitrary edge pruning.
The performance is evaluated using recall, precision, and the percentage of removed edges.

\textbf{Necessity of the Real Geometric Signal.} \texttt{real\_phi} dominates all variants, achieving 87.28\% peak recall and 40.44\% peak precision at $\beta = 0.40$. Conversely, \texttt{uniform\_phi} (no filtering) and \texttt{random\_phi} (random scalar field) stagnate at $\sim$71-72\%. This demonstrates that retrieval gains stem from the learned geometric signal, not merely baseline graph topology (\texttt{uniform\_phi}) or random edge dropout regularization (\texttt{random\_phi}). 

\textbf{Necessity of Topological Alignment.} \texttt{shuffled\_phi} preserves the true numerical distribution of isoperimetric scores but decouples them from graph topology. Consequently, it aggressively removes up to 75\% of edges and suffers a massive performance drop, whereas ISO-RAG achieves peak results by pruning only 32\%. This proves multi-hop retrieval requires topology-aware filtering, not indiscriminate pruning. By strictly aligning the isoperimetric signal with graph structure, ISO-RAG selectively prunes edges leaking probability mass into irrelevant neighborhoods.

\subsection{Qualitative Case Study}

To illustrate how geometric filtering prevents probability leakage in PPR, we examine a multi-hop query from 2WikiMultihopQA (Figure~\ref{fig:phi_casestudy}). This query requires a parallel 2-hop reasoning chain: identifying directors for two films, retrieving their biographies, and comparing their birth dates.

\textbf{Preventing Probability Leakage.} In the top-5 contexts, baseline methods fail to retrieve the crucial passages; their unconstrained PageRank diffusion is hijacked by dense, semantically adjacent hub nodes (e.g., unrelated films or directors). Consequently, HippoRAG2 outputs "Unknown", while HyperbolicRAG hallucinates the wrong film. Conversely, ISO-RAG's isoperimetric control severs spurious edges to these hubs. By bounding probability mass within the local manifold, it retrieves both required passages, enabling the LLM to deduce the correct answer.

\textbf{Decoupling Retrieval and Generation Errors.} To further illustrate the vulnerability of downstream LLMs to contextual noise, we present a "Perfect Retrieval, Failed Generation" case in Figure~\ref{fig:phi_casestudy}. Although ISO-RAG successfully retrieved 100\% of the ground-truth entities, the LLM still failed. This demonstrates that degraded F1/EM scores are not exclusively indicative of retrieval failure, but can also stem from LLM reasoning bottlenecks.

\section{Conclusion}

In this work, we presented ISO-RAG, a novel retrieval-augmented generation framework that imposes strict topological control to mitigate spurious diffusion during graph-based retrieval. By mapping the graph into hyperbolic space and training a geometry-aware encoder, ISO-RAG enables controlled retrieval that is both efficient and effective for downstream question answering. Across benchmarks, it yields consistent  absolute improvements over competitive baselines, underscoring the importance of topology-aware control for multi-hop reasoning.

\bibliography{references}

\end{document}